# Effects of High-Order Co-occurrences on
# Word Semantic Similarity

**Benoît Lemaire**

Laboratoire Leibniz-IMAG (CNRS UMR 5522)

46, avenue Félix Viallet

38031 Grenoble Cedex

France

Benoit.Lemaire@imag.fr

**Guy Denhière**

L.P.C & C.N.R.S. Université de Provence

Case 66, 3 place Victor Hugo

13331 Marseille Cedex, France

denhiere@up.univ-mrs.fr

**Abstract**

A computational model of the construction of word meaning through exposure to texts is built in order to simulate the effects of co-occurrence values on word semantic similarities, paragraph by paragraph. Semantic similarity is here viewed as association. It turns out that the similarity between two words W1 and W2 strongly increases with a co-occurrence, decreases with the occurrence of W1 without W2 or W2 without W1, and slightly increases with high-order co-occurrences. Therefore, operationalizing similarity as a frequency of co-occurrence probably introduces a bias: first, there are cases in which there is similarity without co-occurrence and, second, the frequency of co-occurrence overestimates similarity.

## Introduction

This paper is concerned with semantic similarity. This term is here viewed as association, that is the mental activation of one term when another term is presented, which is what association norms capture. This semantic similarity of two words (or, stated differently, their associative strength) is classically reduced to their frequency of co-occurrence in language: the more frequently two words appears together, the highest is their similarity. This shortcut is used as a quick way of estimating word similarity, for example in order to control the material of an experiment, but it has also an explanatory purpose: people would judge two words as similar because they were exposed to them simultaneously.

The goal of this paper is to study this relation between co-occurrence and similarity by computing similarity and co-occurrence data in a huge corpus of children's texts. Results of our simulation indicate that the frequency of co-occurrence probably overestimates the semantic similarity and that other variables need to be taken into account.

The correlation between co-occurrence and similarity has been found by several researchers (Spence & Owens, 1990). Actually, this relation can be viewed as a simplification of Miller and Charles (1991) hypothesis:

*"two words are semantically similar to the extent that their contextual representations are similar"*

which is usually operationalized into the following assertion, because of computational easiness:

*"two words are semantically similar to the extent that their contextual representations are identical"*

Undoubtly, the frequency of co-occurrence is correlated with human judgement of similarity. However, several researchers have questioned this simple relation. In order to tackle the problem, methodological choices have to be made. People usually restrict their analysis to written texts, although this can be considered as a bias, since we are all exposed to much more language material than just written texts. The first reason is practical: it is much more easier to collect and analyze written texts. In addition, it is probably not a strong bias: if co-occurrence relations in corpora reflect semantic information, they should appear in a similar way in written and spoken languages. The second reason is that, according to Landauer and Dumais (1997), most of the words we know, we learned from texts. The rationale for this assumption is that the spoken vocabulary covers a small part of the whole vocabulary and that direct instruction plays a limited role in word acquisition. This last point is in debate, because the scientific community is lacking of definitive data about how much we are exposed to from texts and from spoken material, and where does our lexical knowledge comes from. However, written material is a good, albeit not perfect, example of the word usage we humans are exposed to. For all of these reasons, co-occurrence analyses are usually performed on written texts.

Studies on large corpora have given examples of words that are strong associates although they never co-occur in paragraphs. For instance, Lund & Burgess (1996) mentionned the two words *road* and *street* that almost never cooccur in their huge corpus although their are almost synonyms. In a 24-million words French corpus from the daily newspaper *Le Monde* in 1999, we found 131 occurrences of *internet*, 94 occurrences of *web*, but no co-occurrences at all. However, both words are strongly associated. The reason why two words are associated in spite of no co-occurrences could be that both co-occur with a third one. For instance, if you mentally construct a new association between *computer* and *quantum* from a set of texts you have read, you will probably construct as well an association between *microprocessor* or *quantum* although they might not co-occur, just because of the existing strong association between *computer* and *microprocessor*. The relation bewween *computer* and *quantum* is called a *second-order co-occurrence*. Psycholinguistic researches on mediated priming have shown that the association between two words can be done through a third one (Livesay & Burgess, 1997; Lowe & McDonald, 2000), even if the reason for that is in debate (Chwilla & Kolk, 2002). Let's go a little further. Suppose that the association between *computer* and *quantum* was also a second-order association, because of another word that co-occured with both words, say *science*. In that case, *microprocessor* and *quantum* are said to be third-order co-occurring elements. In the same way, we can define $4^{th}$-order co-occurrences, $5^{th}$-order co-occurrences, etc. Kontostathis and Pottenger (2002) analyzed such connectivity paths in several corpora and found the existence of these high-order co-occurrences.

However, the question is to know whether these high-order co-occurrences play an important role or not in the construction of word similarities. The answer is not easy since considering only direct co-occurrences sometimes provides good results. In particular, Turney (2001) defines a method for estimating word similarity based on Church and Hanks (1990) pointwise mutual information. The mutual information between x and y is defined as the comparison between the probability of observing x and y together and observing them independently:

$$I(x,y) = \log \frac{P(x,y)}{P(x).P(y)}$$

By extension, this model provides a way to measure the degree of co-occurrence of two words, by comparing the number of co-occurrences to the number of individual occurrences. This value is maximal when all occurrences are co-occurrences. Turney (2001) applied this method to the biggest corpus ever, namely the world wide web. He defined the similarity between two words as the ratio between the number of pages containing both words and the product of the number of pages containing individual occurrences[1]. Turney's similarity is therefore solely based on direct co-occurrences. Turney tested his method using the classical Landauer and Dumais' (1997) TOEFL test: it is composed of 80 items, each containing a stem word and four alternative words from which the participant has to find the closest similar to the stem. Turney applied his method to the test and obtained a score of 73.75%, which is one of the best score ever obtained on this test by a computer without any human intervention.

French and Labiouse (2002) addressed a severe critique on Turney's approach. In particular, they think that this score is high because of stylistic constraints when writing texts: we tend not to repeat words for the sake of style, which explains why synonyms co-occur. Moreover, several works have shown that, although direct co-occurrence gives good results for detecting synonymy, second-order co-occurrence leads to better results. Edmonds (1997) showed that selecting the best typical synonym requires that at least second-order co-occurrence is taken into account. It is true that synonymy can be explained by direct co-occurrence, but second-order co-occurrences probably enhance the relation. In addition, semantic similarity is much more general than pure synonymy. Perfetti (1998) also provides arguments for the weaknesses of direct co-occurrence analyses.

An ideal method would consist in collecting all of the texts subjects have been exposed to and comparing their judgement of similarity with the co-occurrence parameters of these texts. It is obviously impossible. One could think of a more controlled experiment, by asking participants to complete similarity tests before and after text exposure. The problem is that the mental construction of similarities through reading is a long term cognitive process which would probably be invisible over a short period. It also possible to count co-occurrences on representative corpora, but that would give only a global indication a posteriori. This would tell us nothing on the direct effect of a given first or second-order co-occurrence on the semantic similarity. It is valuable to precisely know the effect of direct and high-order co-occurrences during word acquisition. Assume a person X who has been exposed to a huge set of texts since she can read. Let S be the judgement of similarity of X between words W1 and W2. The questions we are interested in are:

- what is the effect on S of X reading a passage containing W1 but not W2?

- what is the effect on S of X reading a passage containing W1 and W2?

- what is the effect on S of X reading a passage containing neither W1 nor W2, but words co-occurring with W1 and W2 (second-order co-occurrence)?

- what is the effect on S of X reading a passage containing neither W1 nor W2, but third-order co-occurring words?

---
[1] He made some variations on the method but the idea is still the same.

Our method is to rely on a model of the construction of word meaning from the exposure to texts in order to trace the construction of similarities according to the occurrence parameters. This model takes texts as input and returns word similarities. It should be cognitively plausible for both inputs and outputs: first, the amount of input texts should be coherent with the quantity of written material people are exposed to and second, the measure of similarity between words should correspond to the human judgment of semantic similarities. For all these reasons, we relied on the Latent Semantic Analysis (LSA) model of word meaning acquisition and representation.

## The cognitive model

### Latent Semantic Analysis

LSA is not only a cognitive model of the *representation* of word meanings but also of its *construction* from the exposure to texts (Landauer & Dumais, 1997). LSA takes as input a large corpus of texts and, after determining the statistical context in which each word occurs, represents each word meaning as a high-dimensional vector, usually composed of several hundreds of dimensions. As opposed to complex symbolic structures, a vector representation is very appropriate for comparing objects since it is straightforward to define a similarity measure. The cosine is an usual measure for that: the highest the cosine, the better the similarity.

Semantic information can indeed be found in raw texts, though in a *latent* form. This is what allows children to understand progressively the meaning of many words by coming across them in various contexts while reading. In LSA, the unit of context is the paragraph. Therefore, LSA first counts the number of occurrences of each word in each paragraph. Words are then represented as vectors. For instance, if the corpus contains 100,000 paragraphs, the word *tree* may be given the following representation, composed of 100,000 numbers: 00102000000......00000. It means that *tree* occurs once in the third paragraph, twice in the fifth, etc. However, this representation is very noisy and dependent on the writers' idiosyncrasies. LSA reduces this huge information in order to only keep the outstanding information. The previous vectors are then represented in an occurrence matrix, from which singular values are extracted. Basically, singular values represent the strength of the previous dimensions. By zeroing the lowest singular values, LSA rules out the noisy and idiosyncratic part of the data. Usually, only a few hundreds dimensions are kept. Tests have shown that performances are maximal around 300 dimensions for the whole language (Landauer et al., 1998), but this value can be smaller when a limited domain is used (Dumais, 2003). Each word is thus represented as a 300-dimensional vector. This high-dimensional space allows a differentiated way of representing polysemic words: the vector corresponding to a unique orthographic form can represent a certain meaning along some dimensions and another one along others, although the dimensions are not labelled at all. For instance, the ambiguous form *fly* is associated to both *plane* (cosine=.48) and *insect* (cosine=.26) but *plane* and *insect* are not associated (cosine=.02) in the "General reading up to 1st year college" semantic space available from the university of Colorado (http://lsa.colorado.edu).

Another interesting point concerns the compositionality of the representation: it is straightforward to go from words to expressions. An expression is given a vector by linear combination of its words. Therefore, the semantic similarity of two expressions can be computed. For instance, using the previous semantic space, the cosine between the two sentences *the cat was lost in the forest* and *my little feline disappeared in the trees* is .37, although they do not share any words except functional words.

Other cognitive models of word meaning representation and acquisition could have been used for our purpose, but none of them fulfilled three important criteria. The first one concerns the input: in order to build a realistic model of children semantic memory development, the input should be of comparable size and nature to what children are exposed to. Models that are based on gigantic corpora could not be used. The second criteria has to do with the output: the model should have semantic similarity results that are similar to those of children in various tasks. The third criteria concerns the model operationalization: our goal is to trace the similarity evolution according to the different kinds of co-occurrences on a large scale, which prevents the use of pure theoretical models or models requiring human intervention. We now detail these three criteria.

**The input criteria**

The input is the nature and size of texts which will be provided to the model. The goal is to reproduce as good as possible what a child is exposed to. First, we will discuss the quantity, then the nature of texts. It is very hard to estimate how many words we process every day. However, we do not need a precise value, but rather a rough idea of the total exposure: is it about a million words, ten millions, a hundred millions? Consider a 20 years-old human, which is approximately the age of participants in psychology experiments. Assume this person reads about one hour a day (this is probably more after the age of 15, but much less before 10). If the reading speed is about 100 words per minute (this is also an average), we end up with a total exposure of 40 millions words. In a similar estimate, Landauer and Dumais (1997) have come to 3,500 words a day, that is 25 millions words at the age of 20. The magnitude is similar to ours. Therefore, we consider a relevant corpus size of tens of million words for adults, and several million words for children around 10.

Hyperspace Analogue to Language (Burgess, 1998) is an interesting model of human semantic memory, but it is currently based on a 300 million words corpus, which is probably overestimated. We do not know however if this amount of input could be reduced whitout altering the model performances. We presented earlier PMI-IR (Turney, 2001); it has very good synonymy measures, but it has been tested on a gigantic corpus, the all web, which is cognitively unplausible.

The nature of the input is also very important. HAL relies on a corpus composed of messages found on Usenet newsgroups. This kind of texts might reproduce exposure to spoken language, but not written material, especially in the case of children. The web pages used by PMI-IR do not either well correspond to what children are exposed to.

We gathered French texts that correspond approximately to what a child is exposed to: stories and tales for children (~1,6 million words), children productions (~800,000 words), reading textbooks (~400,000 words) and children encyclopedia (~400,000 words). This corpus is composed of 57,878 paragraphs for a total of 3.2 million word occurrences. All punctuation signs were ruled out, capital letters were transformed into lower cases, dashes were ruled out except when forming a composed word (like *tire-bouchon*). This corpus was analyzed by LSA and the occurrence matrix reduced to 400 dimensions, which appear to be an optimal value as we will see later. The resulting semantic space contains 40,588 different words.

**The output criteria**

The output of the model is a high-dimensional semantic space, in which the meaning of all words has been represented as vectors. We performed tests in order to check whether these similarities approximately correspond to the children judgement of association. These tests will be presented quickly since they are described in detail elsewhere (Denhière & Lemaire, 2004).

The first test involves a vocabulary task. Material consists of 120 questions, each one composed of a word and 4 definitions: the correct one, a close definition, a far definition and an unrelated definition. This task was performed by 4 groups of children from 2$^{nd}$ grade to 5$^{th}$ grade. These data were compared with the cosines between the given word and each of the four definitions.

Figure 1 displays the percentage of correct answers which is .53 for both the model and the 2$^{nd}$ grade children. The model data follows the same kind of pattern than children data.

Figure 1: Percentage of answers for different kind of definitions

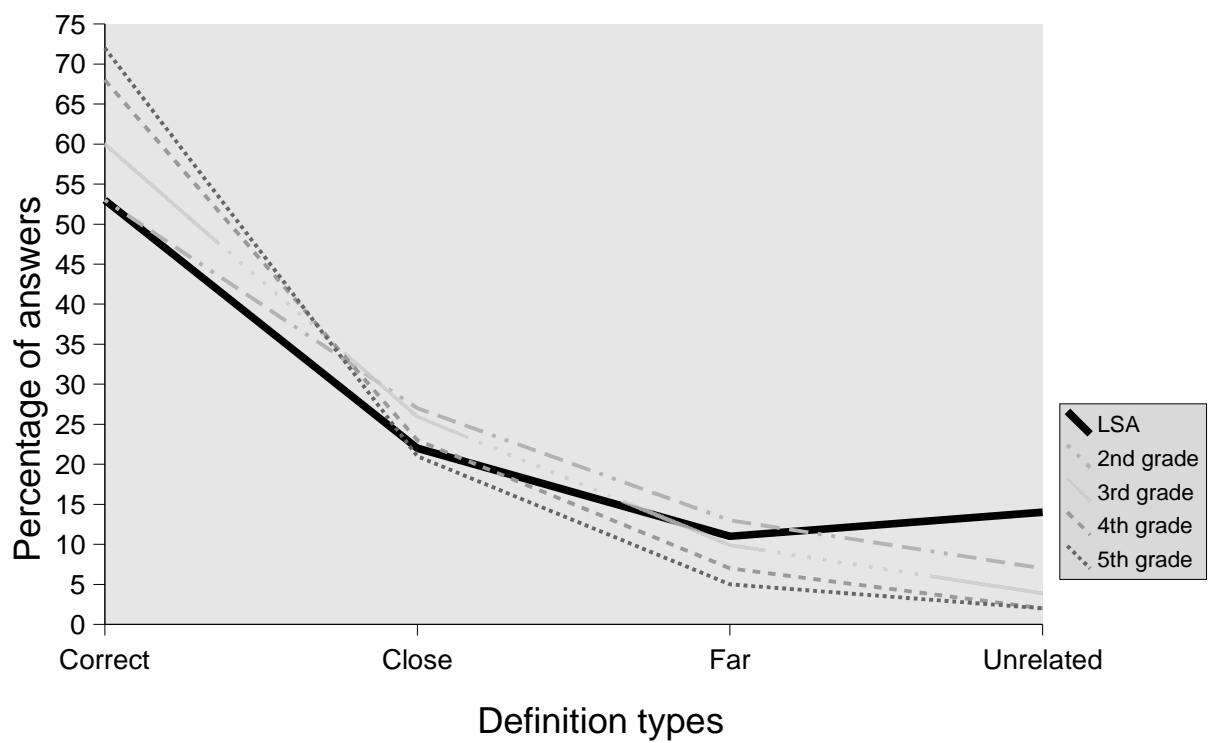

The second experiment is based on verbal association norms published by de la Haye (2003). Two-hundred inducing words were proposed to 9 years-old to 11 years-old children. For each word, participants had to provide the most associated word. The result is a list of words, ranked by frequency. These association values were compared with the LSA cosine between word vectors: we selected the three best-ranked words as well as the three worst-ranked (like in the previous example). We then measured the cosines between the inducing word and the best ranked, the 2nd best-ranked, the 3rd best ranked, and the mean cosine between the inducing word and the 3 worst-ranked. Results are presented in Table 1.

Table 1: Mean cosine between inducing word and various associated words for 9-years-old children

| Words | Mean cosine with inducing word |
|---|---|
| Best-ranked words | .26 |
| 2nd best-ranked words | .23 |
| 3rd best ranked-words | .19 |
| 3 worst-ranked words | .11 |

Student tests show that all differences are significant (p<.03). Our semantic space is not only able to distinguish between the strong and weak associates, but discriminates the first-ranked from the second-ranked and the latter from the third-ranked. Correlation with human data is also significant (r(1184)=.39, p<.001) and raises to .57 when only the 20% most frequent words were considered.

The results of these two tests lead us to consider our children semantic space as a reasonable approximation of the children semantic memory. This is coherent with many researches which have shown that the LSA cosine well mimic the human judgement of semantic association (Foltz, 1996; Landauer, 2002; Wolfe et al., 1998). It is now possible to study in details the effects on the semantic similarity of the different kinds of co-occurrence.

## Simulation

This simulation aims at following the evolution of the semantic similarities of 28 pairs of words over a large number of paragraphs, according to the occurrence values. We started with a corpus size of 2,000 paragraphs. We added one paragraph, ran LSA on this 2001-paragraph corpus and, for each pair, computed the gain (positive or negative) of semantic similarity due to the new paragraph and checked whether there were occurrences, direct co-occurrences or high-order co-occurrences of the two words in the new paragraph. Then we added another paragraph, ran LSA on the 2002-paragraph corpus, etc. Each new paragraph was just the following one in the original corpus. More precisely, for each pair X-Y, we put each new paragraph into exactly one of the following categories:

– occurrence of X but not Y;
– occurrence of Y but not X;
– direct co-occurrence of X and Y;
– second-order co-occurrence of X and Y, defined as the presence of at least three words which co-occur at least once with both X and Y in the current corpus;
– three-or-more-order co-occurrence, which is the rest (no occurrence of X or Y, no direct co-occurrence, no second-order co-occurrence). This category represents three-or-more co-occurrences because paragraphs whose words are completely neutral with X and Y (that is they are not linked to them with a high-order co-occurrence relation) do not modify the X-Y semantic similarity.

We stopped the computation at the 13,637th paragraph. 11,637 paragraphs were thus traced. This experiment took three weeks of computation on a 2 GHz computer with 1.5 Gb RAM. Figure 2 describes the evolution of similarity for the two words *acheter* (*buy*) and *magasin* (*shop*). This similarity is -.07 at paragraph 2000 and raises to .51 at paragraph 13,637. The curve is quite irregular: there are some sudden increases and decreases. Our goal is to identify the reason for these variations.

acheter (buy) / magasin (shop)

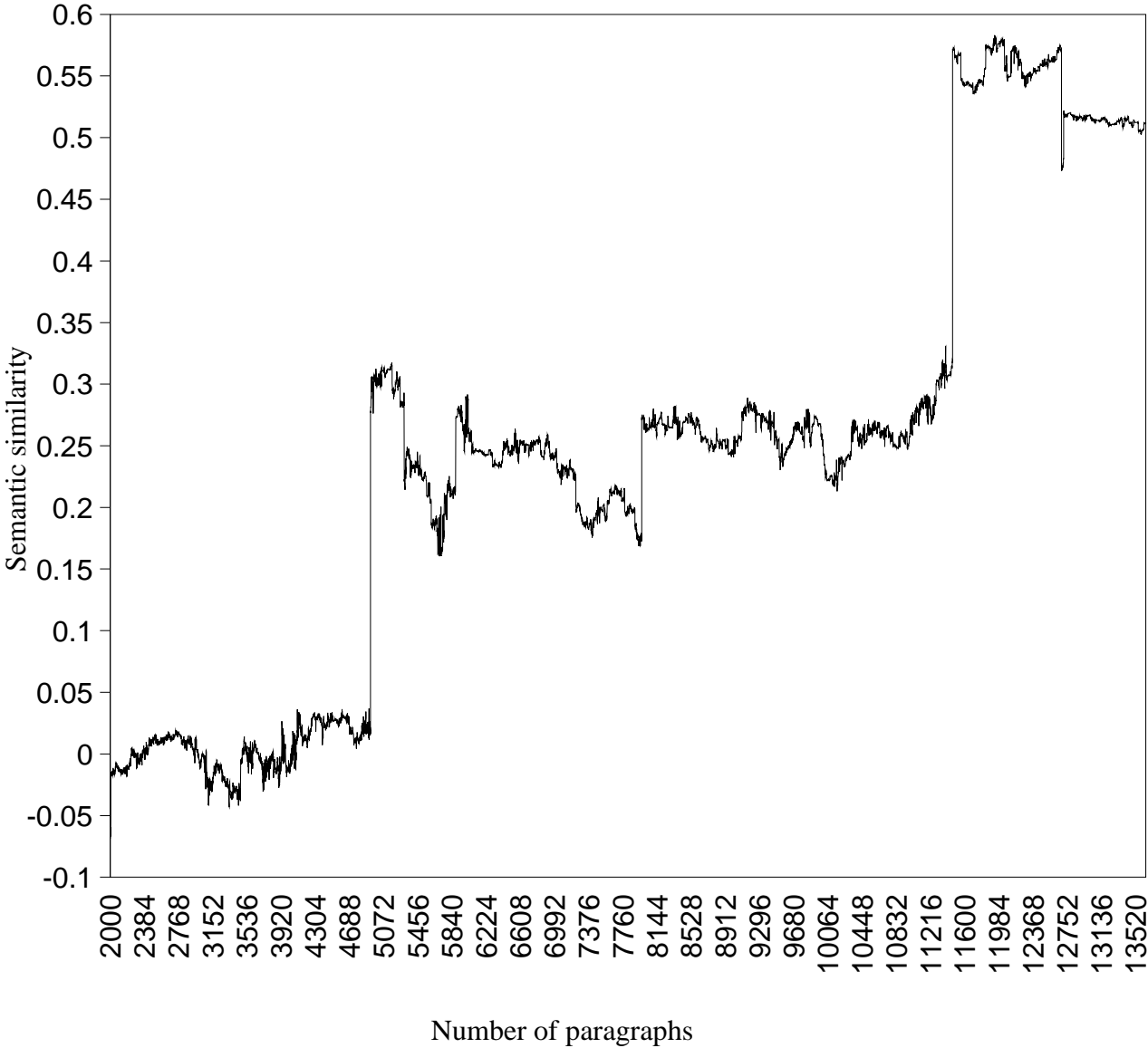

Number of paragraphs

For each pair of word, we shared out the gains of similarity among the different categories. For instance, if the similarity between X and Y was .134 for the 5,000-paragraph corpus and .157 for the 5,001-paragraph corpus, we attributed the .023 gain of similarity to one of the five previous categories. We then summed up all gain for each category. Since the sum of the 11,637 gains of similarity is exactly the difference between the last similarity and the first one, we ended up with a distribution of the total gain of similarity among all categories. For instance, for the pair *acheter (buy)-magasin(shop)*, the .58 (.51 -(-.07)) total gain of similarity is share out in the following way:

– -.10 due to occurrences of only *acheter(buy)*;

– -.19 due to occurrences of only *magasin(shop)*;

– .73 due to the co-occurrences;

– .11 due to second-order co-occurrences;

– .03 due to third-or-more-order co-occurrences.

It means that the paragraphs containing only *acheter(buy)* contributed all together to a decrease of similarity of .10. This is probably due to the fact that these occurrences occur in a context which is different to the *magasin(shop)* context. In the same way, occurrences of *magasin(shop)* led to a decrease of the overall similarity. Co-occurrences tend to increase the similarity, which is expected, and high-order co-occurrences contributes also to an increase.

We performed the same measurement for all 28 pairs of words (Table 2). These pairs were selected from the 200 items of the association task presented earlier and their first-ranked associated word, as provided by children. We only kept words that appear at least once in the first 2,000 paragraphs, in order to have the same number of semantic similarities for all pairs. Table 2 displays the gains of similarities according to the five previous categories for each of the 28 pairs. The first thing is that we found pairs of words that never co-occur (*farine(flour)-gâteau(cake)*) although their semantic similarity increases. Another result is that, except in a few cases, the gain of similarity due to a co-occurrence is higher than the total gain of similarity: in the average, the total gain of similarity is .13 whereas the gain due to a co-occurrence is .34. This is because of a decrease due to occurrences of only one of the two words (-.15 and .-19). In addition, high-order co-occurrences play a small but significant role: they tend to increase the similarity (.14 in total).

Table 2: Distribution of similarity gain among occurrence and co-occurrence categories

| W1-W2 pairs | W1-W2 pairs (translation) | Gain of similarity | Due to | | | | |
|---|---|---|---|---|---|---|---|
| | | | Occurrence of W1 | Occurrence of W2 | Co-occurrence | 2$^{nd}$-order co-occurrence | 3$^{rd}$-order co-occurrence |
| abeille/miel | bee/honey | -.35 | .04 | .27 | .00 | .00 | -.03 |
| acheter/magasin | buy/shop | .58 | -.11 | -.19 | .73 | .11 | .03 |
| avion/vole | plane/fly | .49 | -.26 | -.69 | 1.64 | -.07 | -.14 |
| ballon/jouer | ball/play | .21 | -.15 | -.19 | .32 | .00 | .24 |
| bruit/crier | noise/shout | .13 | .05 | -.02 | .08 | .00 | .02 |
| café/boire | coffee/drink | -.01 | -.07 | -.05 | .03 | .02 | .05 |
| cartable/école | satchel/school | .30 | -.15 | -.20 | .26 | .07 | .32 |
| cave/vin | cellar/wine | .05 | -.05 | .00 | .14 | .05 | -.09 |
| classe/école | class/school | .05 | -.31 | -.34 | .47 | .18 | .04 |
| farine/gâteau | flour/cake | .15 | .09 | -.09 | .00 | .11 | .03 |
| fête/anniversaire | party/birthday | .24 | -.29 | -.25 | .50 | .16 | .12 |
| fourchette/mange | fork/eat | .14 | -.15 | -.17 | .26 | .05 | .15 |
| gentil/méchant | kind/nasty | .36 | -.23 | -.09 | .60 | -.22 | .29 |
| journal/lire | newspaper/rea | -.11 | -.16 | -.11 | .11 | -.16 | .22 |
| lecture/lire | reading/read | .03 | -.21 | -.44 | .61 | -.01 | .08 |
| maman/papa | mummy/daddy | .33 | -.30 | -.35 | .81 | .12 | .05 |
| moulin/farine | mill/flour | -.58 | -.51 | -.30 | .26 | .04 | -.07 |
| noël/cadeau | christmas/gift | .09 | -.14 | -.43 | .37 | -.03 | .31 |
| nuage/ciel | cloud/sky | .10 | -.45 | -.32 | .28 | .11 | .49 |
| parapluie/pluie | umbrella/rain | .20 | -.09 | -.04 | .21 | .15 | -.03 |
| poire/fruit | pear/fruit | .04 | .04 | .01 | .00 | .02 | -.03 |
| rat/souris | rat/mouse | .30 | .05 | -.06 | .20 | -.06 | .17 |
| regarder/voir | watch/see | .16 | -.25 | .05 | .34 | .00 | .02 |
| riche/argent | rich/money | .28 | -.01 | -.11 | .35 | .14 | -.01 |
| rose/fleur | rose/flower | .33 | -.13 | -.11 | .27 | .20 | .10 |
| sable/mer | sand/sea | -.06 | -.35 | -.42 | .31 | .28 | .12 |
| solide/dur | solid/hard | .03 | -.07 | -.16 | .18 | .00 | .08 |
| zèbre/animal | zebra/animal | .10 | -.09 | .13 | .08 | .00 | -.02 |
| AVERAGE | | .13 | -.15 | -.19 | .34 | .05 | .09 |

## Conclusion

It is worth noting that data is quite heterogeneous, but we should not expect all words to follow the same pattern of co-occurrence relations in the language. However, results would have been probably more precise if we could have run the simulation for the whole corpus. In fact, the computational cost of such a simulation is very high, due to the non-incremental behavior of LSA. Moreover, such a cost is not linear: it takes more and more time to process the corpus while paragraphs are added.

Our simulation shows that, although semantic similarity is largely associated to co-occurrence, which is coherent with the literature, the latter overestimates the former. High-order co-occurrences need to be taken into account. By only considering the frequency of co-occurrence as a measure of the semantic similarity, people probably introduce a bias.

## Acknowledgments

This work was done while the first author was in sabbatical at the university of Aix-Marseille. We would like to thank D. Chesnet, E. Lambert and M.-A. Schelstraete, for providing us with parts of the corpus, F. de la Haye for the association data as well as M. Bourguet and H. Thomas for the design of the vocabulary test. We also thank P. Dessus for his comments on a previous version of this paper.